\documentclass{article}

\usepackage{
    dcase2023,
    amsmath,
    graphicx,
    url,
    times,
    booktabs,
    tabularx,
    hyperref,
    color,
    multirow,
    arydshln,
    subcaption,
}

\newcommand\first[1]{\textbf{#1}}  
\newcommand\second[1]{\underline{\smash{#1}}}  
\newcommand\colname[1]{\textbf{#1}}  
\newcommand\example[1]{``\textit{#1}"}
\newcommand\na{\raisebox{0.02cm}{\scriptsize N/A}}

\DeclareMathOperator*{\argmin}{argmin}

\title{Killing two birds with one stone: can an audio captioning system also be used for audio-text retrieval?}

\name{
    Étienne Labbé$^{1}$,
    Thomas Pellegrini$^{1,2}$,
    Julien Pinquier$^{1}$
}
\address{
    $^1$IRIT, Université Paul Sabatier, CNRS, Toulouse, France \\
    $^2$Artificial and Natural Intelligence Toulouse Institute (ANITI) \\
    \{etienne.labbe,thomas.pellegrini,julien.pinquier\}@irit.fr \\
}

\begin{document}

\ninept
\maketitle

\begin{sloppy}

\begin{abstract}

Automated Audio Captioning (AAC) aims to develop systems capable of describing an audio recording using a textual sentence. In contrast, Audio-Text Retrieval (ATR) systems seek to find the best matching audio recording(s) for a given textual query (Text-to-Audio) or vice versa (Audio-to-Text). These tasks require different types of systems: AAC employs a sequence-to-sequence model, while ATR utilizes a ranking model that compares audio and text representations within a shared projection subspace. However, this work investigates the relationship between AAC and ATR by exploring the ATR capabilities of an unmodified AAC system, without fine-tuning for the new task. Our AAC system consists of an audio encoder (ConvNeXt-Tiny) trained on AudioSet for audio tagging, and a transformer decoder responsible for generating sentences. For AAC, it achieves a high SPIDEr-FL score of 0.298 on Clotho and 0.472 on AudioCaps on average. For ATR, we propose using the standard Cross-Entropy loss values obtained for any audio/caption pair. Experimental results on the Clotho and AudioCaps datasets demonstrate decent recall values using this simple approach. For instance, we obtained a Text-to-Audio R@1 value of 0.382 for AudioCaps, which is above the current state-of-the-art method without external data. Interestingly, we observe that normalizing the loss values was necessary for Audio-to-Text retrieval.
\end{abstract}


\section{Introduction}
\label{sec_intro}
In recent years, audio-language tasks have received greater attention due to advances in machine learning for text processing. For example, the Automated Audio Captioning (AAC) task aims to create machine learning systems that produce a sentence describing an audio file, while the Audio-Text Retrieval (ATR) task aims to use a caption to extract an audio from its database (Text-to-Audio, T2A) or use an audio to retrieve a caption from its database (Audio-to-Text, A2T). Research on these tasks is also boosted by the DCASE Challenge and Workshop\footnote{\url{https://dcase.community/}}, which proposed two tasks dedicated to AAC and T2A. Although these tasks appear to be closely related, they are usually performed by two different systems and architectures. Those systems can sometimes share common weights~\cite{xu2022_t6a}, but they need to be trained differently on several phases. In the image captioning task, the authors of~\cite{9616897} proposed to use a captioning system by describing each image and compare these descriptions to the captions instead of the images. In this paper, we propose another method for using an AAC system to perform the ATR task, and we investigate the implications of using this system in this way.

\section{System description}

\subsection{AAC system architecture}
To achieve the AAC task, we employ a deep neural network with an encoder-decoder architecture. We trained a ConvNeXt~\cite{liu2022convnet} (CNext) model for audio tagging and used it as an encoder to produce frame-level features to overcome the lack of audio-language data. The ConvNeXt was trained on the AudioSet~\cite{audioset} audio tagging dataset without the AudioCaps~\cite{kim_etal_2019_audiocaps} audio captioning dataset files to avoid biases. This encoder achieves a high mAP score of 0.462 on AudioSet. The details of the architecture and training hyperparameters are given in~\cite{pellegrini2023adapting}. The encoder gives a list of features of shape $768 \times 31$ for a 10-seconds audio clip, which are projected by a sequence of dropout set to 0.5, dense layer, a ReLU activation and another dropout set to 0.5. The decoder is a standard transformer decoder architecture~\cite{NIPS2017_3f5ee243} with six decoder layers blocks, four attention heads per block, a feedforward dimension of 2048, a GELU~\cite{gelu} activation function and a global dropout set to 0.2. Unlike a lot of AAC and ATR systems, no pre-trained weight has been used for the decoder/word modelling part. We found that freezing the ConvNeXt encoder leads to lower variances, so we decided to pre-compute all its embeddings to train only the decoder part. The whole model contains 28M frozen parameters and 12M trainable parameters.

\subsection{Data augmentation}
During our training with the decoders, we added three different augmentations on audio and input word embeddings to reduce overfitting and improve model generalization. Mixup~\cite{zhang2018mixup} modifies the input audio and word embeddings during training, with $\alpha$ set to 0.4. Each embedding is mixed with another one in the current batch, except for the target, which remains unmixed. Label Smoothing~\cite{szegedy2015rethinking} is applied to the target one-hot vectors to reduce the maximal probability of each word and limit the confidence of the model. Finally, SpecAugment~\cite{Park_2019} masks a part of the audio frame embeddings, with 6 stripes dropped with a maximal size of 4 in time axis and 2 stripes dropped with a maximal size of 2 in feature axis.

\begin{table*}[!htbp]
    \centering
    \caption{AAC results on Clotho and AudioCaps testing subsets. Our results are averaged over 5 seeds. WC stands for WavCaps~\cite{mei2023WavCaps} dataset. Best values for each dataset/metric are in \first{bold}, and best values without external data are \second{underlined}.}
    \begin{tabular}{@{}clccccccc@{}}
        \toprule
        \colname{Dataset} & \colname{System} & \colname{Train data} & \colname{\#params} & \colname{METEOR} & \colname{CIDEr-D} & \colname{SPICE} & \colname{SPIDEr} & \colname{SPIDEr-FL} \\
        \midrule
        \multirow{3}{*}{CL}
        & BEATs+Conformer~\cite{wu2023_t6a} & CL+AC & {127M} & \first{.193} & \first{.506} & \first{.146} & \first{.326} & \first{.326} \\
        \noalign{\vskip 0.01cm}
        \cdashline{2-9}
        \noalign{\vskip 0.1cm}
        & CNN14-trans~\cite{won2021_t6} & CL & {88M} & {.177} & {.441} & {.128} & {.285} & \na \\
        & CNext-trans (ours) & CL & {40M} & \second{.189} & \second{.464} & \second{.136} & \second{.300} & \second{.298} \\
        \midrule
        \multirow{3}{*}{AC}
        & HTSAT-BART~\cite{mei2023WavCaps} & {AC+WC} & {171M} & \first{.250} & \first{.787} & {.182} & \first{.485} & \na \\
        \noalign{\vskip 0.01cm}
        \cdashline{2-9}
        \noalign{\vskip 0.1cm}
        & Multi-TTA~\cite{kim2023exploring} & {AC} & {108M} & {.242} & \second{.769} & {.181} & \second{.475} & \na \\
        & CNext-trans (ours) & AC & {40M} & \second{.246} & {.763} & \first{\second{.183}} & {.473} & {.472} \\
        \bottomrule
    \end{tabular}

    \label{tab_captioning_results}
\end{table*}

\subsection{Using a captioning system for retrieval}
The first idea to use an AAC system for ATR is to generate predictions to describe each audio file and compare each text query to each description using a metric like BLEU, CIDEr-D or SBERT, as proposed in~\cite{9616897}, but we found low results using this strategy. We believe that AAC systems tend to produce less detailed and diversified sentences than references, which leads to a loss of information when using it to summarize the audio content into a single sentence. Typically, the vocabulary size used during inference is only around 617 distinct words over the 1839 words present on average in the references for the Clotho development-testing subset.
AAC systems are usually trained to predict the next token of a sentence using previous words and the audio file. This means that the model actually takes as input an audio and a caption, and the loss could be used to score this input. We decided to simply use the Cross-Entropy (CE) loss used in training to score each pair, and expecting that an AAC system should be able to give a higher loss value when the input caption does not match the input audio file than when they match. Equations~\ref{eq_t2a} and \ref{eq_a2t} describe how an audio and text element are retrieved using the CE.

\begin{subequations}
\begin{equation}
    \text{T2A}(t, A, f) = \argmin_{a \in A} \text{CE}(f(a, t_{\text{prev}}), t_{\text{next}})
    \label{eq_t2a}
\end{equation}
\begin{equation}
    \text{A2T}(a, T, f) = \argmin_{t \in T} \text{CE}(f(a, t_{\text{prev}}), t_{\text{next}})
    \label{eq_a2t}
\end{equation}
\end{subequations}

\noindent
where $t$ corresponds to a caption, $T$ is the list of all captions, $a$ is an audio file from the $A$ list of audio files. $f$ is the AAC system which produces the distributions of probabilities for the next words $t_{next}$ given the previous words $t_{prev}$ in the context of an audio file.

\section{Experimental setup}

\subsection{Datasets}
\label{sec_datasets}


AudioSet~\cite{audioset} is the largest audio tagging dataset publicly available and contains 2M pairs of audio/tag. The audio files last for 10 seconds extracted from YouTube videos and the dataset contains 527 different sound events tags. Clotho~\cite{drossos_clotho_2019} (CL) is an AAC dataset containing 6974 audio files ranging from 15 to 30 seconds in length extracted from the FreeSound website. The dataset is divided into three splits used respectively for training, validation and testing, containing five captions per audio file. In our experiments, each audio file is resampled from 44.1 kHz to 32 kHz. During training, we randomly select one of five captions for each audio file. AudioCaps~\cite{kim_etal_2019_audiocaps} (AC) is the largest AAC dataset written only by humans, containing 51308 audio files from the AudioSet dataset. Since original YouTube videos are removed or unavailable for various reasons, our version of the train split contains 46230 out of 49838 files, 464 out of 495 in the validation split and 912 out of 975 files in the test split. In addition, we slightly improve caption correctness in the training subset by manually fixing 996 invalid captions with grammatical and typographic errors. For the two AAC datasets, captions are put in lowercase and all punctuation characters are removed. The codebase used to download, read and extract data is a package named aac-datasets\footnote{\url{https://pypi.org/project/aac-datasets/0.3.3/}}.


\subsection{Metrics}


For the AAC task, we report the five metrics used in the DCASE Challenge task 6a. METEOR~\cite{denkowski_meteor_2014} is based on the precision and recall of the words. CIDEr-D~\cite{vedantam_cider_2015} uses the TF-IDF scores of the shared n-grams between candidates and references. SPICE~\cite{anderson_spice_2016} builds a graph representing the scene described by the captions and computes an F-score with its common edges. SPIDEr~\cite{liu_improved_2017} averages the two previous metrics and finally, SPIDEr-FL\footnote{\url{https://dcase.community/challenge2023/task-automated-audio-captioning}} is a combination of the SPIDEr metric with a pre-trained system designed to detect fluency errors. When one of them is detected, the SPIDEr score is divided by a factor of 10. The codebase for AAC metrics is available as a public Pip package\footnote{\url{https://pypi.org/project/aac-metrics/0.4.2/}} named aac-metrics. For the ATR task, we use the Recall@k metric, which measures if a relevant (ground truth) element is in the top-k retrieved elements.

\begin{table*}[htbp]
    \centering
    \caption{Audio-language retrieval results on Clotho and AudioCaps testing subsets. Our results are averaged over five seeds. WC stands for WavCaps dataset. Best values for each dataset/task/metric are in \first{bold}, and best values without external data are \second{underlined}. The asterisk * denotes the results scaled by a min-max strategy described in~\ref{ssec_a2t_perf_low}.}
    \begin{tabular}{@{}clcc|ccc|ccc@{}}
        \toprule
        {\colname{Retrieval}} & \multirow{2}{*}{\colname{System}} & {\colname{Training}} & \multirow{2}{*}{\colname{\#params}} & \multicolumn{3}{c}{\colname{Text-to-audio}} & \multicolumn{3}{c}{\colname{Audio-to-text}} \\
        \colname{dataset} && {\colname{dataset(s)}} & &  \colname{R@1} & \colname{R@5} & \colname{R@10} &  \colname{R@1} & \colname{R@5} & \colname{R@10} \\
        \midrule
        \multirow{6}{*}{CL}
        & PaSST-N$^4$~\cite{primus2023_t6b} & CL+AC+WC & {441M} &  \first{.261} & \first{.553} & \first{.693} & \na & \na & \na \\
        & CNN14-BERT~\cite{mei2023WavCaps} & CL+WC & {214M} & {.215} & {.479} & {.663} &  \first{.271} & \first{.527} & \first{.663} \\
        \noalign{\vskip 0.01cm}
        \cdashline{2-10}
        \noalign{\vskip 0.1cm}
        & CNN14-BERT~\cite{wang2023_t6b} & {CL} & {192M} & {.167} & \second{.410} & \second{.539} & \na & \na & \na \\
        & Triplet-weighted~\cite{mei2022metric} & {CL} & {185M} & {.142} & {.366} & {.497} & {.169} & {.381} & {.514} \\
        & TAP+PMR~\cite{xin2023improving} & {CL} & {185M} & \second{.171} & {.396} & \na & \second{.182} & {.399} & \na \\
        & CNext-trans (ours) & {CL} & {40M} & {.137} & {.349} & {.480} & \hphantom{a}{.148}* & \hphantom{a}\second{.404}* & \hphantom{a}\second{.541}* \\
        \midrule
        \multirow{7}{*}{AC}
        & HTSAT-BERT~\cite{mei2023WavCaps} & {AC+WC} & {141M} & {.422} & {.765} & {.871} & \first{.546} & \first{.852} & \first{.924} \\
        & ONE-PEACE~\cite{wang2023onepeace} & {CL+AC+7 others} & {2B} & \first{.425} & \first{.775} & \first{.884} & {.510} & {.819} & {.920} \\
        \noalign{\vskip 0.01cm}
        \cdashline{2-10}
        \noalign{\vskip 0.1cm}
        & MMT~\cite{Koepke_2022} & {AC} & {290M} & {.361} & {.720} & {.845} & {.396} & {.768} & {.867} \\
        & Multi-TTA~\cite{kim2023exploring} & {AC} & {187M} & {.347} & {.703} & {.832} &  {.402} & {.740} & {.872} \\
        & TAP+PMR~\cite{xin2023improving} & {AC} & {185M} & {.368} & {.727} & \na & {.417} & {.762} & \na \\
        & CNN14+TAP+PMR~\cite{xin2023improving} & {AC} & {192M} & {.334} & {.688} & \na & \second{.431} & {.733} & \na \\
        & CNext-trans (ours) & {AC} & {40M} & \second{.382} & \second{.733} & \second{.853} & \hphantom{a}{.398}* & \hphantom{a}\second{.814}* & \hphantom{a}\second{.919}* \\
        \bottomrule
    \end{tabular}

    \label{tab_retrieval_results}
\end{table*}

\subsection{Hyperparameters}

The number of training epochs $K$ is set to 400 with a batch size set to 512. The optimizer used is AdamW with an initial learning rate ($\text{lr}_0$) set to $5 \cdot 10^{-4}$, $\beta_1$ set to $0.9$, $\beta_2$ set to {$0.999$}, $\epsilon$ set to $10^{-8}$ and weight decay set to {$2$}. Weight decay is not applied to the bias contained in the network. The learning rate is decreased during training at the end of each epoch $k$ using a cosine scheduler rule: $\text{lr}_k = \frac{1}{2} \big(1 + \cos ( \frac{k \pi}{K} ) \big) \text{lr}_0$. The gradient $L_2$-norm is clipped to $1$ to avoid collapsing across seeds, the label smoothing reduces maximal target probability by 0.2 and the mixup $\alpha$ hyperparameter is set to 0.4. Since only the projection and the decoder part are trained, a single AAC experiment runs in one hour on AC and three hours on CL datasets with one V100 graphics card. To validate our model, we used the FENSE metric~\cite{fense} which is based on the cosine similarity of the embeddings produced by a pre-trained Sentence-BERT model combined with the same fluency error detector used in SPIDEr-FL. During validation and inference, we used the standard beam search algorithm to generate better sentences. In order to limit the number of repetition tokens, we forced the model to avoid generating the same word twice in a single sentence, except for stop words defined in the NLTK package.



\section{Results}

\subsection{AAC and ATR results}
The AAC results are given in Table~\ref{tab_captioning_results}. We also reported the SOTA scores for each dataset, without reinforcement learning, without ensemble method and with or without external captioning datasets. On CL, our model performs better than the previous SOTA without external data (CNN14-trans) in all metrics and uses more than twice fewer parameters (40M instead of 88M). We believe this is mainly due to our stronger pretrained encoder, which has a higher mAP score on AudioSet and produces better features for AAC. On AC, the model reach a score very close to the Multi-TTA method, with only 0.002 absolute difference in SPIDEr despite having an unbiased encoder not trained on the testing files of AC.

Retrieval results are shown in Table~\ref{tab_retrieval_results}. Just as AAC results, we reported the SOTA methods without ensemble methods and with or without external captioning datasets. Since all values are not always reported, we added several SOTA methods to compare our system with at least one other methods for each column. For the T2A task on the CL dataset, our model performs better than the DCASE baseline, but worse than most SOTA methods. However, the system achieves the highest scores on AudioCaps without external data. Somewhat surprisingly, our system outperforms other methods without external data on the A2T task on R@5 and R@10, but not on the R@1 metric on both datasets.

\begin{figure*}
    \centering
    \hfill
    \begin{subfigure}{0.33\textwidth}
        \includegraphics[width=1.0\textwidth]{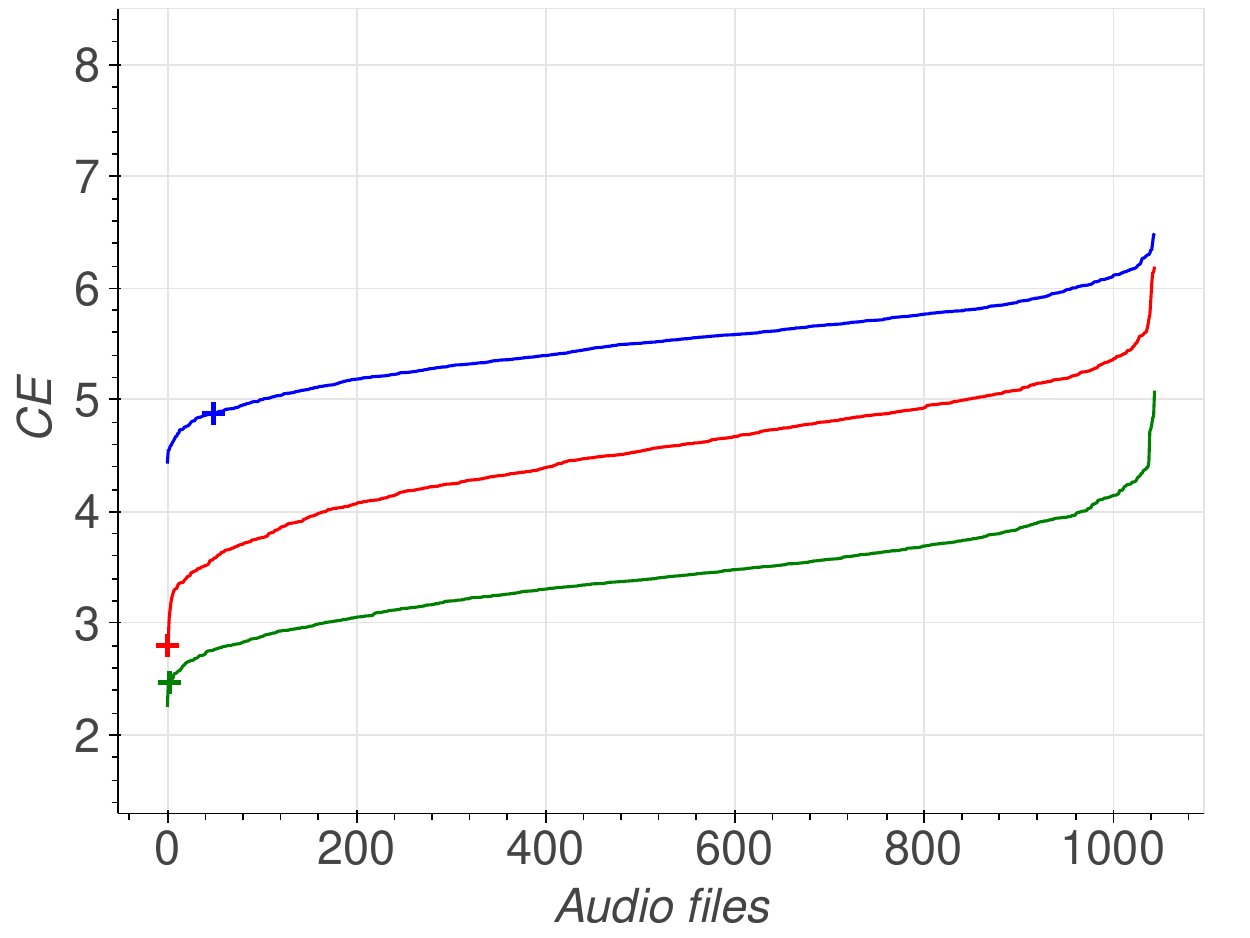}
        \caption{Sorted losses for 3 captions over audios.}
        \label{sfig_t2a}
    \end{subfigure}
    \hfill
    \begin{subfigure}{0.33\textwidth}
        \includegraphics[width=1.0\textwidth]{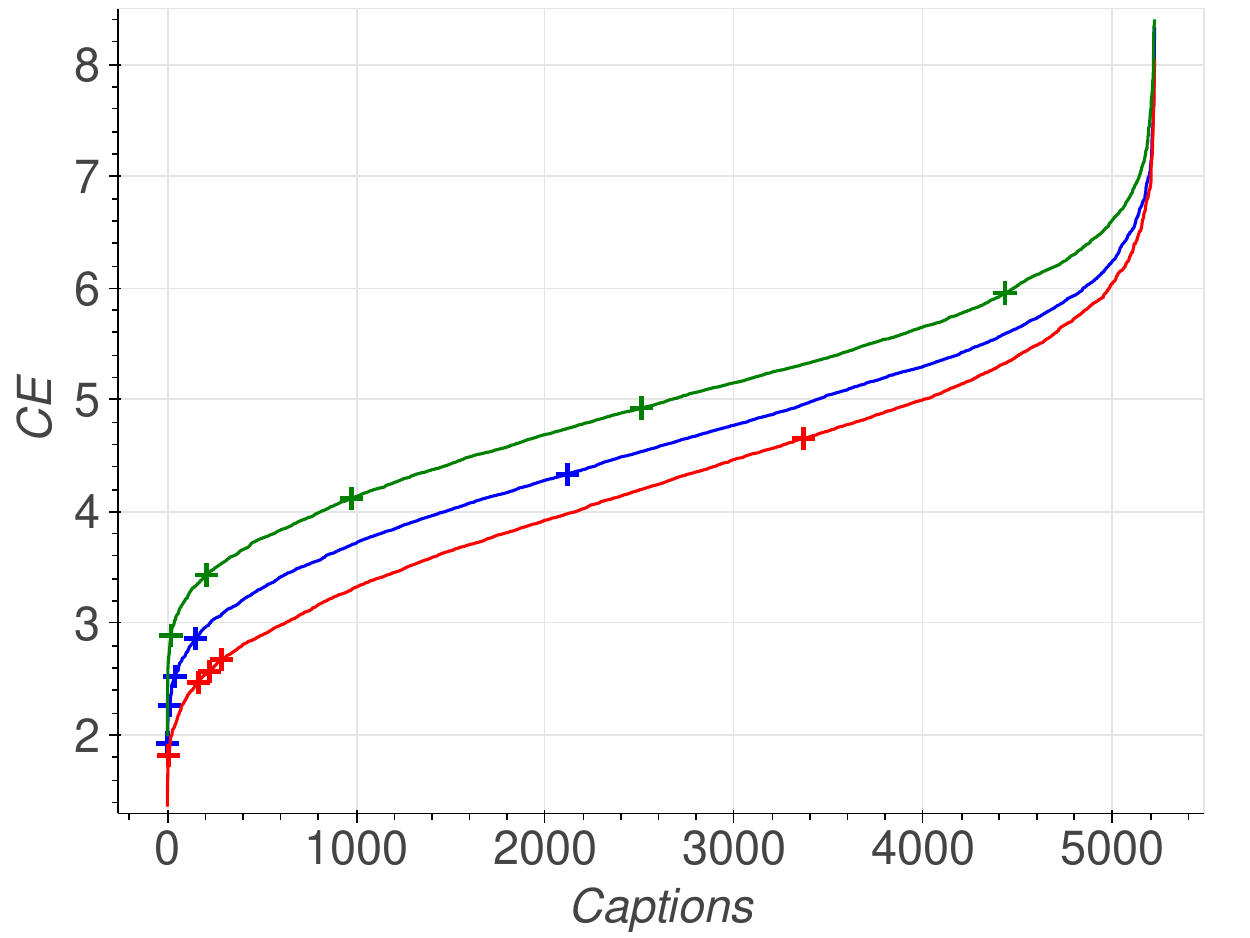}
        \caption{Sorted losses for 3 audios over captions.}
        \label{sfig_a2t}
    \end{subfigure}
    \hfill
    \begin{subfigure}{0.33\textwidth}
        \includegraphics[width=1.0\textwidth]{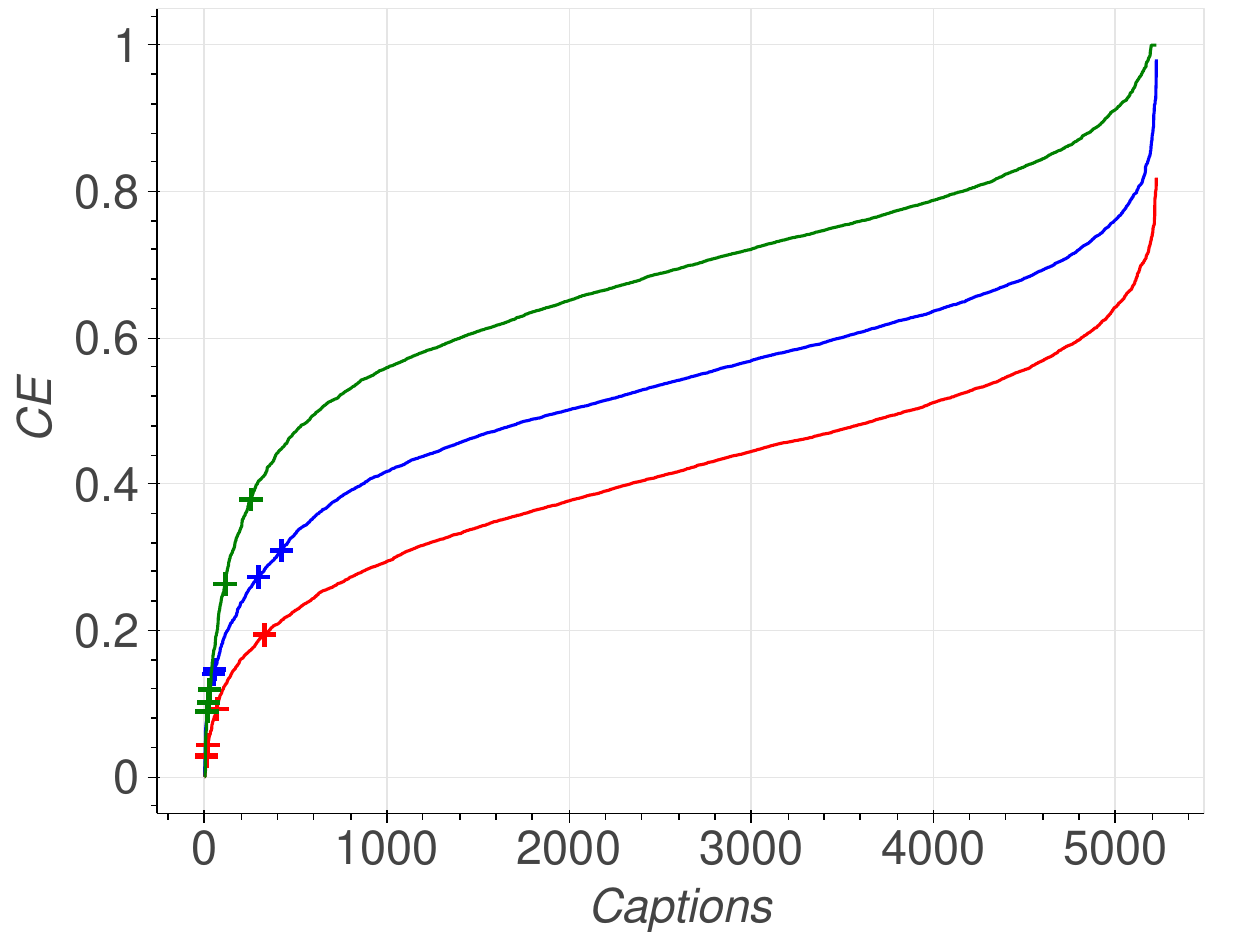}
        \caption{\textbf{Scaled} sorted losses for 3 audios over captions.}
        \label{sfig_a2t_scaled}
    \end{subfigure}
    \hfill
    \caption{Losses for 3 queries over all retrieved items. The position of the relevant (ground truth) elements are shown with a cross.}
    \label{fig_losses}
\end{figure*}

\subsection{A2T min-max scaling}
\label{ssec_a2t_perf_low}

We found that even if our system performs well on T2A task, the results on A2T one were really low compared to the SOTA ones. The system reaches an R@1 of 0.146 on AC and 0.038 on CL when using raw loss values. We found that this is caused by a subset of the captions, where the loss values are almost always lower than the others for all audio files. For instance, in Figure~\ref{sfig_t2a}, the vertically lowest green curve corresponds to the loss of a query with all the other audio files, and is almost always lower than the other curves. In particular, only 120 unique captions are retrieved for 1045 queries during the A2T task with raw losses, but we did not find a strong correlation between these captions and the frequencies of their words or their length. In order to clarify why it impacts only the A2T task and not T2A, we provide a simple example in Table~\ref{tab_toy_ex}. This example shows the loss values for three different audio A$_i$ with their corresponding captions C$_i$. When we perform the T2A task, we select the retrieved audio A$_i$ with the lowest loss value in the column $i$, which achieves a perfect score in that case. However, when we perform A2T, only the caption C$_1$ is retrieved, because its column has a range of value different from the others, which explains the poor results when using raw loss values. To tackle this problem, we propose a post-processing which scales each ``column" (i.e., each series of values corresponding to a single caption). In particular, we tried to normalize and standardize, but a simple min-max scaling has led to the best results. We also added a rule when two retrieved captions has the same score (zero when they are the minimal value of their column) by using their original losses to decide which one will be used. The impact of this scaling on the A2T losses are given in figures~\ref{sfig_a2t} and~\ref{sfig_a2t_scaled}.

\begin{table}[htbp]
    \centering
    \caption{Real loss values over 3 audio files and captions.}
    \begin{tabular}{c|ccc}
        \hline
        & C$_1$ & C$_2$ & C$_3$ \\
        \hline
        A$_1$ & \first{1.7} & 8.4 & 8.1 \\
        A$_2$ & 2.1 & \first{7.6} & 8.5 \\
        A$_3$ & 2.0 & 8.3 & \first{6.5} \\
        \hline
    \end{tabular}
    \label{tab_toy_ex}
\end{table}

\vspace{-4mm}
\section{Benefits and downsides of using AAC system}

Recently, the authors of paper~\cite{wu2023audiotext} showed that ATR systems usually fail to capture high-level relations between sounds by showing corrupted captions to an ATR system. More precisely, they propose to replace in caption the word \example{after} by \example{before} and vice versa to invert the sequence of sound events described and name this the Before-After Test (BAT). The ATR system should be able to give a lower score for an incorrect input caption than for a correct one. We believe that audio-language systems should be able to capture that kind of information better than audio event classes, but the actual metrics do not usually reflect the model performance on it. In addition to the perturbation proposed by them, we proposed to switch the relation type from sequence to superposition and vice versa by replacing some words or inverting the propositions of the sentence. For example, the sentence \example{a man speaks \textbf{then} a dog barks} can become \example{a man speaks \textbf{as} a dog barks} if we replace \example{then}, or become \example{a dog barks \textbf{then} a man speaks} if we invert the propositions between \example{then}. We detailed the different words tested in Table~\ref{tab_repl_inv_sets}.


\begin{table}[htbp]
    \centering
    \caption{Detailed words used for Replace. BAT stands for Before-After-Test, seq2sup for sequence-to-superposition and sup2seq for superposition-to-sequence.}
    \begin{tabular}{c|c|c}
        \toprule
        \colname{Set} & \colname{Words} & \colname{Replaced by one of} \\
        \midrule
        \multirow{2}{*}{BAT} & before & after \\
        & after & before \\
        \midrule
        \multirow{2}{*}{seq2sup} & followed by, and then, & \multirow{2}{*}{as, while} \\
        & then, before, after & \\
        \midrule
        \multirow{2}{*}{sup2seq} & \multirow{2}{*}{as, while} & followed by, and then, \\
        & & then, before, after \\
        \bottomrule
    \end{tabular}
    \label{tab_repl_inv_sets}
\end{table}
\vspace{-5mm}

\begin{table}[htbp]
    \centering
    \caption{Accuracy over different perturbations on Clotho development-testing subset. 0.5 is the score of a random model.}
    \begin{tabular}{lccc}
        \toprule
        \colname{System} & \colname{Type} & \colname{Set} & \colname{Accuracy} \\
        \midrule
        MLP~\cite{wu2023audiotext} & \multirow{4}{*}{Replace} & \multirow{4}{*}{BAT} & {.496} \\
        MLP+ACBA~\cite{wu2023audiotext} & & & {.554} \\
        TFMER~\cite{wu2023audiotext} & & & {.509} \\
        TFMER+ACBA~\cite{wu2023audiotext} & & & {.685} \\
        \noalign{\vskip 0.01cm}
        \hdashline
        \noalign{\vskip 0.1cm}
        \multirow{3}{*}{CNext-trans (ours)} & \multirow{3}{*}{Replace} & BAT & {.768} \\
        & & seq2sup & {.825} \\
        & & sup2seq & {.903} \\
        \noalign{\vskip 0.01cm}
        \hdashline
        \noalign{\vskip 0.1cm}
        \multirow{3}{*}{CNext-trans (ours)} & \multirow{3}{*}{Invert} & BAT & {.892} \\
        & & seq & {.906} \\
        & & sup & {.778} \\     
        \bottomrule
    \end{tabular}
    \label{tab_relation_tests}
\end{table}

The Table~\ref{tab_relation_tests} shows that our model performs very well at discriminating sound events relations, with 76.8\% for the BAT, higher than the best of the compared study (68.5\%). We can also see that our model performs very well on other tests which perturb the relations, with 90.6\% It could imply that our model effectively captures the sequence and superposition relations. We also noticed for the Invert test with superposition words that our model is still able to detect the correct caption, probably because the first sounds described in those sentences are the loudest or longest ones in the audio.

Nevertheless, an AAC system requires computing the whole decoder pass-forward for each pair audio/caption, while usually ATR systems compute separate embeddings for each modality. For the A2T task, the post-processing is required to achieve an acceptable performance, necessitating to keep the minimal and maximal value of the loss for each caption, or an estimation of them. If a new caption is added to the database, the minimal and maximal value also need to be computed or estimated with several audio files. This scaling should also be required for zero shot experiments, which is close to the A2T task.

\vspace{-1mm}
\section{Conclusions}
In this study, we propose a straightforward method for leveraging any standard AAC system for A2T. We demonstrate that despite not being specifically trained for it, an AAC system can achieve reasonable performance on both the T2A and A2T subtasks. Furthermore, it can even attain state-of-the-art scores compared to ATR methods that do not employ external data. We also observed that our model often overestimates the loss value for a subset of captions in the A2T task, resulting in poor results in the initial configuration. To address this issue, we introduced a post-processing strategy based on min-max scaling to mitigate bias in the scores. This adjustment significantly improved the results, for instance, increasing R@1 from 0.038 to 0.148 on Clotho. Finally, we evaluated our system by perturbing the input captions and found that it outperforms another ATR method in distinguishing various sound event relations. In the future, potential research directions could involve modifying AAC training using a contrastive-based loss to enhance ATR performance or developing new benchmarks and test databases to refine the evaluation of ATR systems.



\vspace{-1mm}
\section{ACKNOWLEDGMENT}
\label{sec_ack}
This work was partially supported by the Agence Nationale de la Recherche LUDAU (Lightly-supervised and Unsupervised Discovery of Audio Units using Deep Learning) project (ANR-18-CE23-0005-01). This work was granted access to the HPC resources of IDRIS under the allocation 2022-AD011013739 made by GENCI.

\bibliographystyle{IEEEtran}
\bibliography{refs}

\end{sloppy}
\end{document}